\documentclass[journal]{IEEEtran}

\IEEEoverridecommandlockouts
\usepackage{cite}
\usepackage{amsmath,amssymb,amsfonts}
\usepackage{graphicx}
\usepackage{textcomp}
\usepackage{xcolor}
\usepackage{algorithm}
\usepackage{algpseudocode}
\usepackage[table]{xcolor}
\usepackage{multirow}
\usepackage{makecell}
\usepackage{soul}
\usepackage{url}
\usepackage[hidelinks]{hyperref}
\usepackage[utf8]{inputenc}
\usepackage[small]{caption}
\usepackage{subcaption}
\usepackage{float}
\usepackage{graphicx}
\usepackage{amsmath}
\usepackage{booktabs}
\usepackage{color}
\usepackage{amssymb}
\def\BibTeX{{\rm B\kern-.05em{\sc i\kern-.025em b}\kern-.08em
    T\kern-.1667em\lower.7ex\hbox{E}\kern-.125emX}}
\begin{document}

\title{Balancing Fairness, Privacy, and Accuracy: A Multitask Adversarial Framework for Centralized Data-Driven Systems\\
\thanks{Code available at: \href{https://anonymous.4open.science/r/PFA-776F/README.md}{anonymous.4open.science }{\url{https://anonymous.4open.science/r/PFA-776F/README.md}}.}
}
\author{
Imesh Ekanayake,
Elham Naghizade,
and Jeffrey Chan%
\thanks{Imesh Ekanayake, Elham Naghizade, and Jeffrey Chan are with the School of Computing Technologies, RMIT University, Melbourne, VIC, Australia.}
\thanks{E-mail: imesh.ekanayake@rmit.edu.au; e.naghizade@rmit.edu.au; jeffrey.chan@rmit.edu.au.}
}

\maketitle

\begin{abstract}
The integration of fairness and privacy in centralized data-driven applications is critical, especially as these systems increasingly influence sectors with significant societal impact. Current methods rarely address privacy, fairness, and accuracy together, which can potentially compromise ethical standards and privacy regulations. However, balancing these three objectives is quite challenging since each of objective often imposes conflicting requirements on the design and training of models, making it difficult to optimize one without compromising the others.
This paper introduces a novel multitask adversarial model that treats fairness and privacy as integral objectives rather than afterthoughts, and learns a latent representation that hides sensitive attributes while preserving essential task-related information.
Our approach dynamically balances fairness with accuracy and privacy through an optimized cost function with minimal performance loss even under strict conditions. Extensive testing on diverse datasets shows the ability of our model to achieve high standards of fairness and privacy without significant sacrifice to accuracy. Benchmarking against state-of-the-art privacy and fairness standards shows that our method enhances the robustness of privacy, fairness, and accuracy optimization, proving its adaptability across various datasets.
\end{abstract}


%%%%%%%%%%%%%%%%%%%%%%%%%%%%%%%%%%%%%%%%%%%%%%%%%%%%%%%%%%%%%%%%%%%%%%%%%%%%%%%%%%%%%%%%%%%%%

\section{Introduction}
AI has significantly impacted virtually every sector, enhancing efficiency and decision-making~\cite{pessach2022review}. However, these developments pose ethical challenges regarding the utilization of sensitive information, highlighting the need for a framework to balance utility, privacy, and fairness.
%However, these developments pose ethical challenges when managing data, highlighting the need for a framework to balance accuracy, privacy, and fairness.

Ensuring algorithmic fairness as well as preserving privacy are well-studied in the context of centralized data driven systems, particularly in healthcare and finance. However, their joint effect on applications remains under-addressed.
For example, during the COVID-19 pandemic, the use of proctoring software in education raised concerns about privacy \cite{lee2022online} and fairness \cite{nigam2021systematic}, with criticisms of invasive surveillance and biases, particularly against students of color\cite{cheuk2021can}. Algorithmic decision-making in various sectors have also raised bias and privacy concerns. The COMPAS system in the judiciary has been criticized for perpetuating racial biases and legal issues, with studies indicating that it might incorrectly predict higher recidivism risks among minorities~\cite{dressel2018accuracy}.

\begin{figure*}[htp]
  \centering
  \begin{subfigure}{0.45\textwidth}
    \centering
    \includegraphics[width=\linewidth]{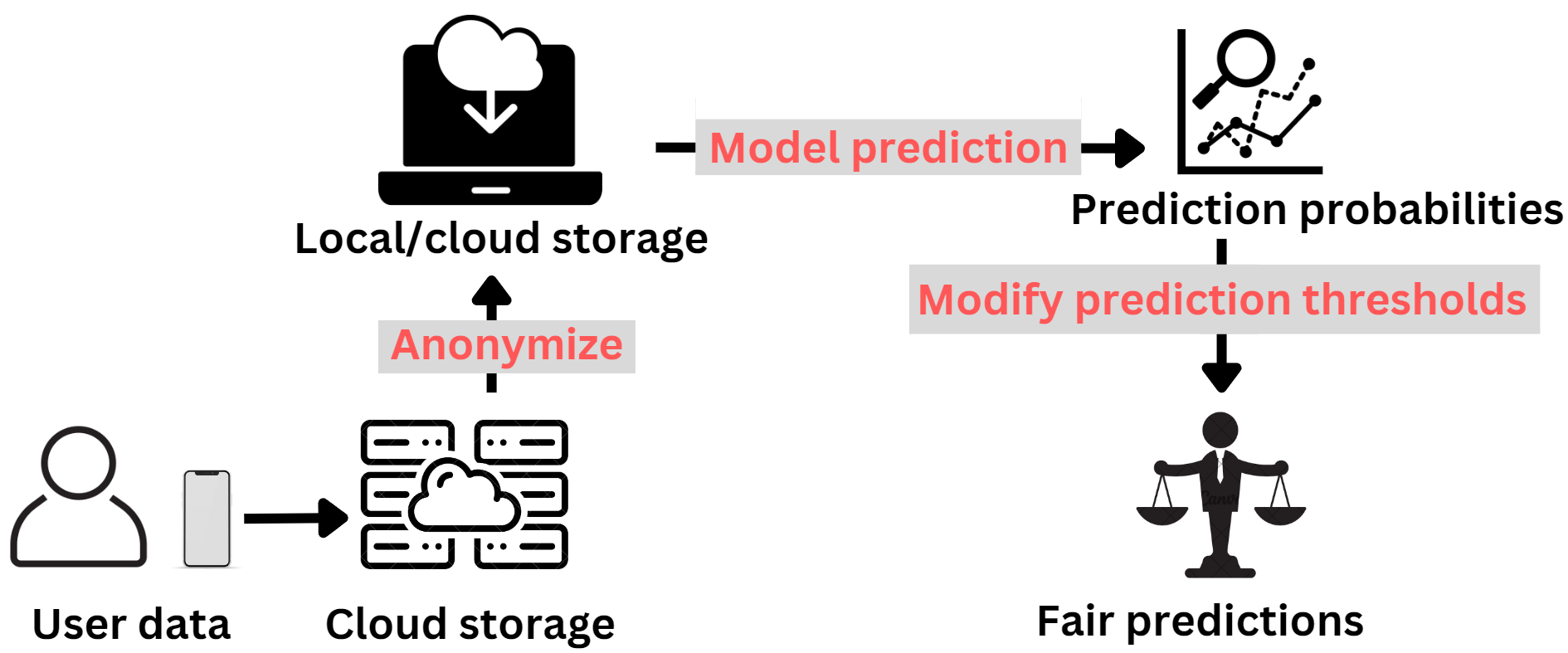}
    \caption{(a) Current standards for ensuring privacy and fairness. The model
generates prediction probabilities based on anonymized data, which are
then subjected to threshold modifications to uphold fairness criteria~\cite{JagielskiKMORSU19,DBLP:journals/corr/abs-2001-04958}.
However, the need for manual threshold modification for each demo-
graphic group indicates a process that is neither private nor robust in
ensuring fairness.}
 
   % Current standards for ensuring privacy and fairness. The model generates prediction probabilities based on anonymized data, which are then subjected to threshold modifications to uphold fairness criteria. However, the need for manual threshold modification for each demographic group indicates a process that is neither private nor robust in ensuring fairness.}
    \label{fig:general_method}
  \end{subfigure}
  \hfill
  \begin{subfigure}{0.45\textwidth}
    \centering
    \includegraphics[width=\linewidth]{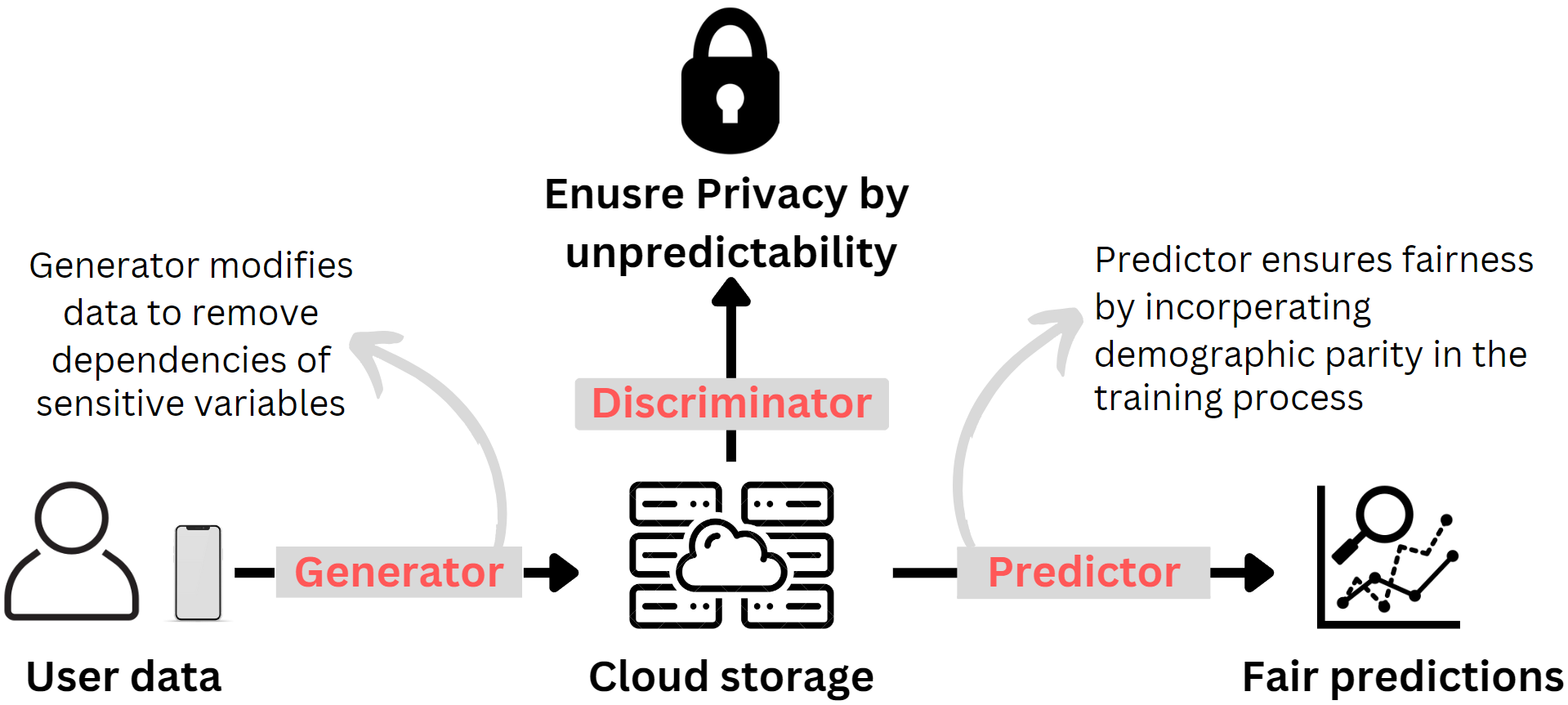}
    \caption{Our proposed framework, which integrates three key components: 
    The Generator modifies user data to remove dependencies on sensitive (private) variables. The Discriminator ensures that the transformations applied by the Generator maintain data utility without revealing sensitive information. Concurrently, the Predictor incorporates fairness into the training process.}
    \label{fig:our_framework}
  \end{subfigure}
  \caption{Ensuring privacy and fairness as separate processes (Figure\ref{fig:general_method}) vs. our proposed multitask adversarial framework  (Figure\ref{fig:our_framework}). }
  \label{fig:combined_fig}
\end{figure*}

% Whilst this urges the design of methods to balance privacy, fairness, and accuracy in ML models, such goal is challenging~\cite{agarwal2021trade} as these objectives are often in conflict, making it difficult to optimize one without compromising the others:
Whilst the necessity of protecting sensitive data urges the design of methods to balance privacy, fairness, and accuracy in ML models, achieving this is challenging \cite{agarwal2021trade}. These objectives are often in conflict, making it difficult to optimize one without compromising the others
Enhancing privacy, typically involves techniques like adding noise to the data (e.g., differential privacy)~\cite{huang2024differential} or transforming features to reduce their identifiability (e.g., feature anonymization)~\cite{ye2024securereid}. While these methods effectively mitigate the risk of data breaches and ensure the privacy of sensitive attributes, they often degrade the quality of the data and consequently a loss in model accuracy because the noise or transformations obscure the signals that models rely on to make predictions.
On the other hand, maintaining fairness often requires altering training procedures to ensure that models do not disproportionately favor or disadvantage any particular group based on sensitive attributes like race, gender, or age~\cite{hu2024enhancing}. Techniques such as re-weighting training examples or modifying output thresholds to balance error rates across groups can lead to a reduction in the overall model accuracy\cite{small2024equalised}. This is because these adjustments force the model to perform equally well across groups, potentially at the expense of missing out on higher accuracy that could be achieved on a less constrained model.
Finally, ensuring both fairness and privacy can be particularly challenging because methods to anonymize data or protect individual identities can remove or alter characteristics that are crucial for achieving fairness~\cite{mittal2024responsible}. For instance, removing or encrypting sensitive attributes to protect privacy can prevent the model from using these attributes to correct for biases, thus impairing the model's ability to enforce fairness constraints.
Figure~\ref{fig:general_method} depicts current standards to ensure privacy and fairness, where each objective is treated as a separate process which often leads to compromised outcomes where enhancing one can detrimentally affect the other.

To mitigate this limitation, we propose a novel multi-task adversarial model that treats privacy and fairness as \emph{primary, co-equal objectives} rather than post-hoc constraints. Our model (Figure~\ref{fig:our_framework}) first learns a latent space in which sensitive attributes are difficult to recover while the task-relevant structure is retained, motivating the ensuing dual-adversarial training strategy. The predictor and synthetic-data generator jointly optimize for accuracy and fairness by embedding a demographic-parity penalty into their cost functions, whereas an adversarial classifier is trained to infer sensitive attributes from the latent representation; the predictor and generator are updated to make this inference unsuccessful, thereby enhancing privacy. We use demographic parity as the primary training constraint because it controls the allocation of beneficial outcomes independently of labels that may encode historical bias, and it matches applications where equal access to positive decisions across demographic groups is required. 

Extending traditional machine learning models beyond the sole optimization of accuracy increased the complexity of the cost function. In the proposed methodology, the overall loss was written as a weighted linear combination of three components corresponding to prediction error, unfairness, and privacy leakage, each scaled by a non-negative coefficient (denoted $\alpha$, $\beta$, and $\gamma$, respectively). While this scalarization followed the standard weighted-sum formulation, the novelty did not lie in the use of a linear combination per se, but in the specific adversarially defined fairness and privacy surrogates and in the way the weights were selected. The coefficients were not hand-tuned; instead, they were treated as outer-loop design variables and were calibrated by a multi-objective search procedure over held-out fairness, privacy, and accuracy metrics, so that none of the criteria dominated and Pareto-efficient operating points were identified. Technically, the difficulty arose from the strong, non-linear coupling between the three objectives through shared network parameters and adversarial training, which made naive weighting schemes unstable. The proposed calibration procedure yielded a robust and flexible optimization framework for classification tasks, validated through extensive empirical evaluations.

%%%%%%%%%%%%%%%%%%%%%%%%%%%%%%%%%%%%%%%%%%%%%%%%%%%%%%%%%%%%%%%%%%%%%%%%%%%%%%%%%%%%%%%%%%%%%

\section{Related Work}

Related work on privacy preservation, fairness enhancement in predictive models, and the resulting accuracy--privacy--fairness trade-off was reviewed in this section. The literature was grouped into fairness-enhanced learning, privacy-preserved learning, and joint fairness--privacy methods.

\subsection{Fairness-Enhanced Machine Learning}

Fairness in machine learning was treated as essential as AI was increasingly used in high-stakes decisions. Methods were typically categorized as pre-processing, in-processing, and post-processing. In pre-processing, training data were adjusted to reduce bias through re-weighting or dataset modification \cite{kamiran2012data,calmon2017optimized}. In in-processing, fairness constraints were embedded in model training \cite{zemel2013learning}. In post-processing, predictions were altered after training to enforce demographic parity or equalized odds \cite{hardt2016equality,pleiss2017fairness,rafi2024fairness}.

In practice, group-fairness criteria were selected by application context and data quality. Demographic parity was often favored under biased historical data, ensuring group-level outcome independence and relevance in lending or justice settings \cite{han2023retiring}, and it was argued to promote long-term equity \cite{small2024equalised}. Equal opportunity and equal odds were constrained by their need for richer labeled data \cite{hardt2016equality}. Adversarial learning was also used to build unbiased predictors while balancing accuracy, highlighting inherent trade-offs \cite{woodworth2017learning}. For unlabeled or label-scarce regimes, unsupervised fairness approaches such as fair clustering and adversarial representations were developed to prevent bias \cite{chierichetti2017fair,kleindessner2019fair,edwards2015censoring}.

\subsection{Privacy-Preserved Machine Learning}

\paragraph{Adversarial privacy.}
Adversarial representation learning was used to hide sensitive attributes by training generators that produced embeddings on which discriminators could not recover protected information beyond chance. This approach was shown to remove gender or race signals in recommender systems without harming personalization \cite{10.1145/3643670}. Similar encoders such as AFR \cite{pmlr-v80-madras18a} and FAIR-VAE \cite{kairouz2022generating} were reported to preserve utility without externally injected noise, provide distribution-adaptive local privacy, and remain compatible with differentiable models, but instability in min--max optimization, added hyperparameters, and the lack of worst-case guarantees were noted.

\paragraph{Differential privacy (DPriv).}
DPriv was defined to bound any single record’s influence \cite{dwork2014algorithmic}. Mechanisms included output perturbation, gradient perturbation via DP-SGD \cite{abadi2016deep}, and teacher--student schemes such as PATE. Strong guarantees were provided, but accuracy was usually reduced because noise lowered the signal-to-noise ratio, requiring larger batches, more epochs, or higher privacy budgets $\varepsilon$ \cite{xu2019achieving}. DPriv noise was also reported to risk widening fairness gaps when minority groups were under-represented.

\paragraph{Bridging the gap.}
Federated learning was used to limit raw-data exposure \cite{yang2023federated}, yet centralized domains requiring unified ledgers were noted to weaken its benefits and introduce drift \cite{truong2021privacy,xu2020federated}. Under such settings, FairDP-SGD and FairPATE were proposed to trace Pareto frontiers over DP, demographic parity, and accuracy \cite{yaghini2023learning}. When $\varepsilon$ was varied from $1$ to $10$, competitive accuracy was retained at moderate budgets, while steep degradation was observed once $\varepsilon<2$.

\subsection{Fairness-Enhanced and Privacy-Preserved Machine Learning}

Joint optimization of utility, demographic parity, and privacy was framed as a three-way trade-off. PFairDP was introduced as a modular pipeline in which fairness enforcement, DPriv training, and utility learning were separated into tunable modules and optimized in a multi-objective manner \cite{ficiu2023pfairdp}. DIR and ROC were used for fairness control \cite{feldman2015disparate,kamiran2012decision}, and DP-SGD parameters were used for privacy control \cite{abadi2016deep}. PFairDP was compared with constraint-guided baselines by Pannekoek and Spigler \cite{pannekoek2021tradeoffs} and Xu et al.\ \cite{xu2019dp_fair_logreg}, and with grid and random search; higher-hypervolume Pareto sets were reported on Adult and MEPS, matching or exceeding earlier constraint-based surfaces \cite{ficiu2023pfairdp}. FairDP (FAIRDP) was proposed to certify group fairness under DPriv by partitioning data by protected groups, training group-specific DP-SGD models, and aggregating them to equalize group contribution, with Monte Carlo inference used to certify demographic-parity and equal-opportunity bounds \cite{tran2023fairdp,abadi2016deep}. Comparisons were made against DP-SGD \cite{abadi2016deep}, DPSGD-F \cite{xu2021removing_disparate_impact_dp}, FairSmooth \cite{jin2022fairsmooth}, DP-IS-SGD \cite{kulynych2022principled}, and DPSGD-Smooth \cite{tran2023fairdp}; improved fairness--utility trade-offs were reported on Adult, Default-CCC, and UTK-Face \cite{tran2023fairdp}.

Additional lines tightened or explained the same trade-off. Wasserstein Differential Privacy was used to reduce leakage with marginal accuracy loss \cite{DBLP:conf/aaai/YangQZ24}. Robust optimization paired with DPriv was shown to shrink group-error gaps at a utility cost \cite{hansen-etal-2024-impact}. Impossibility bounds were derived showing that the trade-off could not be eliminated \cite{li2024limits}. Theory further linked privacy noise to fairness shifts by bounding how mechanisms such as output perturbation and DP-SGD altered group metrics and utility \cite{mangold2023differential}. The fairness impact of privacy noise was shown to shrink with more data or weaker budgets but remain non-negligible in small or imbalanced regimes, motivating strong DP baselines and reporting of privacy--fairness--utility frontiers.

%%%%%%%%%%%%%%%%%%%%%%%%%%%%%%%%%%%%%%%%%%%%%%%%%%%%%%%%%%%%%%%%%%%%%%%%%%%%%%%%%%%%%%%%%%%%%

\section{Proposed PFA Framework}

This study introduces a novel multitask adversarial model designed to balance the three competing objectives, i.e., privacy, fairness and accuracy in classification tasks.

\subsection{Problem Statement}
\label{sec:problem}

\begin{figure*}[h]
  \includegraphics[width=0.88\linewidth]{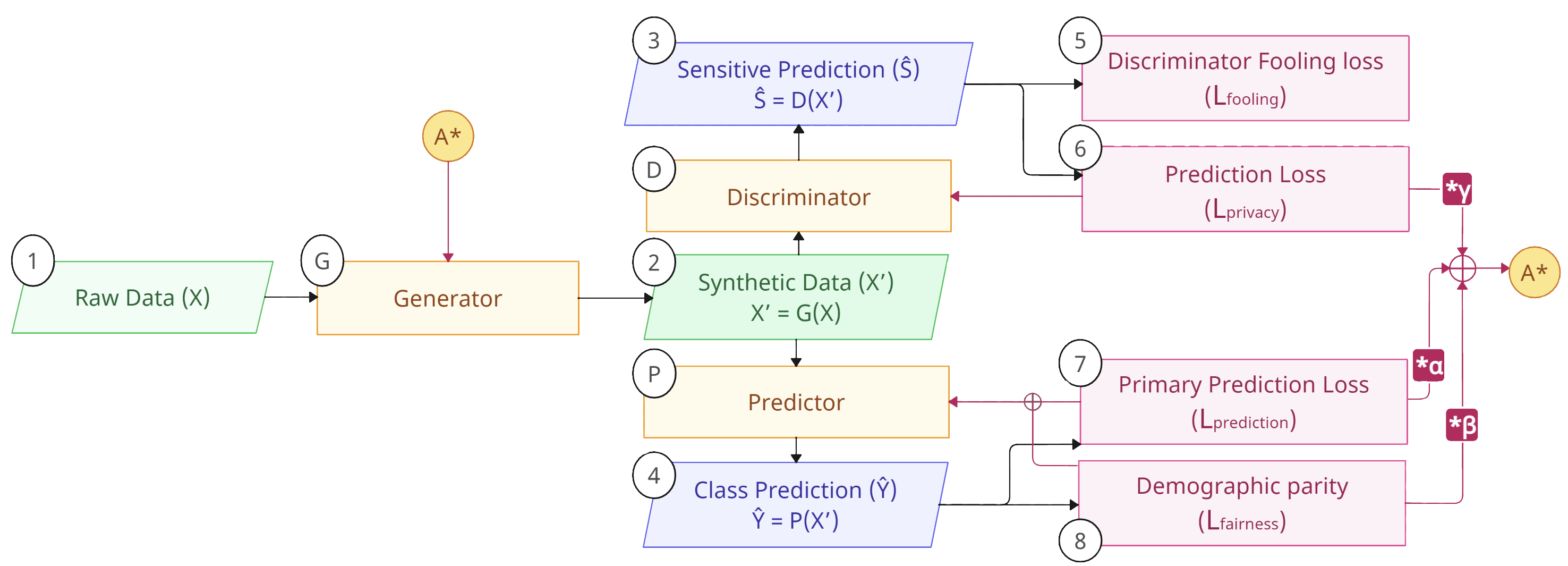}
  \caption{Multi-task adversarial framework for Privacy-Fairness-Accuracy (PFA) 
    integrates \textit{feedforward} (black lines) and \textit{backpropagation} (red lines) flows to concurrently optimize privacy, fairness, and accuracy. Raw data \(X\) enters the Generator \(G\), producing modified data \(X'\). This data feeds into both the Discriminator \(D\) and the Predictor \(P\). The Discriminator outputs \(\hat{S}\), and its discriminator prediction loss \(L_{\text{Privacy}}\) and fooling loss \(L_{\text{fooling}}\) guide \(G\) to better mask sensitive attributes. The Predictor estimates \(\hat{Y}\) and assesses accuracy with prediction loss \(L_{\text{prediction}}\) and fairness with demographic-parity loss \(L_{\text{Fairness}}\). The Generator is optimized using \(L_{\text{fooling}}\), \(L_{\text{prediction}}\), and \(L_{\text{fairness}}\), balanced by cost coefficients \(\alpha\), \(\beta\), and \(\gamma\). This strategy achieves a unified balance among the model's core objectives.}
    
  \label{fig:PFA}
\end{figure*}

% Given a dataset, $\mathcal{D} = \{x_i, s_i, y_i\}_{i=1}^{N}$ where each data point has features $x_i \in \mathcal{X}$, a binary sensitive attribute, i.e., $s_i \in \{0,1\}$ and a binary task label, i.e., $y_i \in \{0,1\}$. We assume $\phi$ is one of the features in $\mathcal{X}$
% that captures whether or not the $i$th record belong to a protected group. We aim to learn a function $f: \mathcal{X} \rightarrow \mathcal{X'}$,  a classifier $g: \mathcal{X'} \rightarrow \{0,1\}$ and a sensitive attribute classifier $h: \mathcal{X'} \rightarrow \{0,1\}$ such that the following objectives are jointly optimized:
Given a dataset $\mathcal{D} = \{x_i, s_i, y_i\}_{i=1}^{N}$ where each data point had features $x_i \in \mathcal{X}$, a binary sensitive attribute $s_i \in \{0,1\}$, and a binary task label $y_i \in \{0,1\}$, a toy loan-approval setting was considered to make the setup concrete. Here, $y_i = 1$ indicated that applicant $i$ repaid a past loan, and $s_i = 1$ denoted membership in a protected group (e.g., female applicants), while $s_i = 0$ denoted the non-protected group. One of the features $\phi \in \mathcal{X}$ captured this group membership (e.g., $\phi(x_i) = 1$ if $i$ belonged to the protected group and $0$ otherwise).
In this example, a function \(f: \mathcal{X} \rightarrow \mathcal{X}'\) is learned to map raw applicant features (income, credit score, and employment history) into a latent representation \(z_i = f(x_i)\). A classifier \(g: \mathcal{X}' \rightarrow \{0,1\}\) is then trained to predict loan approval decisions \(\hat{y}_i = g(z_i)\), and a sensitive-attribute classifier \(h: \mathcal{X}' \rightarrow \{0,1\}\) is trained to predict \(\hat{s}_i = h(z_i)\). Intuitively, \(g\) is encouraged to preserve information in \(z_i\) that is useful for predicting \(y_i\), while \(h\) is used adversarially so that \(z_i\) contains as little information as possible about \(s_i\). In the loan example, this means that two applicants with similar financial profiles but different group membership are encouraged to receive similar approval scores, thereby promoting demographic parity while still using informative, non-sensitive features.

\color{black}

\begin{enumerate}
    \item Accuracy of the classification task is maximized through minimizing the classification error, i.e.: $$
\min_{f,g} \mathbb{E}_{(x, y) \sim \mathcal{D}} \left[ \mathcal{L}_{\text{prediction}}(g(f(x)), y) \right]
$$

    \item Privacy is maximized through maximizing the confusion of the sensitive attribute classifier, i.e.:
    $$
\max_{f}\min_{h} \mathbb{E}_{(x, s) \sim \mathcal{D}} \left[ \mathcal{L}_{\text{privacy}}(h(f(x)), s) \right]
$$

\item Fairness is maximized through minimizing demographic parity. Defining demographic parity, $\mathcal{F}_{DP}(g, f, \mathcal{D})$ as bellow, the aim is to $ \min_{f,g} \mathcal{F}_{DP}(g, f, \mathcal{D})$. $$\mathcal{F}_{DP}(g, f, \mathcal{D}) = |P(g(f(x))|\phi = 0) - P(|g(f(x))|\phi = 1)|$$ 
\end{enumerate}

Our proposed \textbf{P}rivacy-\textbf{F}airness-\textbf{A}ccuracy (PFA) framework incorporates the following essential components and processes to meet the above objectives by treating privacy, fairness, and accuracy as three \emph{competing, strongly coupled} objectives defined on a shared representation and by forcing them to be optimized \emph{simultaneously} within a generator–predictor–discriminator architecture. Optimizing three separate models for utility, fairness, and privacy and combining them afterwards would allow each objective to pull the representation in incompatible directions; in contrast, PFA constrains a single generator to mediate all three losses at once, making the trade-offs explicit in one learned latent space:

\begin{enumerate}
    \item \textbf{Input Variables}: 
    The model inputs raw user $i$th data for $1 \leq i \leq N$, comprising features \(x_i\) and the sensitive attribute \(s_i\), to predict outcome \(y_i\).
    % while ensuring privacy and fairness.

    \item \textbf{ Generator}:
    The generator \(G\) transforms input data \(\mathcal{X}\) into \(\mathcal{X}'\) to obscure the sensitive attribute \(s\), preserving data utility for predicting \(y\) and promoting fairness. The modifications prevent the inference of sensitive information, which preserves privacy.
    
    \item \textbf{Predictor}: 
    The predictor \(P\) uses transformed data \(X'\) for the primary task of predicting \(y \in \{0,1\}\), while incorporating fairness via demographic parity in its cost function to balance prediction rates across demographic groups.

    \item \textbf{Discriminator}: 
    The discriminator \(D\) predicts the sensitive attribute \(s \in \{0,1\}\) from transformed data \(X'\). The adversarial relationship with the generator enhances data privacy by making \(X'\) difficult for \(D\) to predict.

    \item \textbf{Optimization Process}:
    The PFA model's optimization targets a unified objective function balancing three elements: Privacy Loss \(L_{\text{privacy}}\) (lower values indicate better protection of sensitive attributes by the discriminator), Fairness Loss \(L_{\text{fairness}}\) (ensuring equitable treatment across groups using demographic parity), and Prediction Loss \(L_{\text{prediction}}\) (accuracy of outcome predictions). This method effectively harmonizes privacy, fairness, and accuracy.

\end{enumerate}

The PFA framework optimizes privacy, fairness, and accuracy objectives simultaneously as detailed in Section~\ref{sec:cost}.

\subsection{Loss Functions}
\label{sec:cost}

The model employs three main loss functions associated with the Generator, Discriminator, and Predictor, respectively. 

\subsubsection{Predictor Loss}
The Predictor loss,$\mathcal{L}_{\text{P}}$, combines the primary prediction loss and the fairness loss:

\begin{equation}
\label{eq:prediction_loss}
\mathcal{L}_{\text{prediction}} = \frac{1}{N} \sum_{i=1}^{N} \ell_{\text{BCE}} (y_i, \hat{y}_i)
\end{equation}

where 
\begin{equation}
\label{eq:bce_loss}
\ell_{\text{BCE}} = - \left[ y_i \log(\hat{y}_i)  + (1 - y_i) \log(1 - \hat{y}_i) \right]
\end{equation}

\begin{equation*} 
\mathcal{L}_{\text{fairness}} = 
% L_{\text{DP}} =
\left| \frac{1}{N_0} \sum_{i \in S_0} \hat{y}_i - \frac{1}{N_1} \sum_{i \in S_1} \hat{y}_i \right| 
\end{equation*}
where $S_0$ and $S_1$ correspond to the subset of data points such that $s_i = 0$ and $s_i = 1$ respectively. $N_0$ and $N_1$ also correspond to the cardinality of 

\begin{equation} 
\mathcal{L}_{\text{P}} = \mathcal{L}_{\text{prediction}} +   \mathcal{L}_{\text{fairness}} 
\end{equation}

\subsubsection{Privacy Loss}
The Discriminator employs the Binary Cross Entropy loss function for predicting the sensitive variable:
, where \(\) represents the sensitive attribute, and \(\hat{S}\) is its predicted probability.

\begin{equation}
\mathcal{L}_{\text{privacy}} = \frac{1}{N} \sum_{i=1}^{N} \ell_{\text{BCE}} (s_i, \hat{s}_i)
\end{equation}

\subsubsection{Generator Loss}

As stated in Section~\ref{sec:problem}, we need to maximize the confusion of the sensitive attribute classifier. This is captured through the Discriminator's fooling cost, which is also modeled as a Binary Cross Entropy loss aimed at misleading the Discriminator to improve data privacy:
\begin{equation} 
\mathcal{L}_{\text{fooling}} = \frac{1}{N} \sum_{i=1}^{N} - \left[ \log(D(G(x_i))) + \log(1 - D(G(x_i)))\right] 
\end{equation}

The Generator is optimized with a composite objective that simultaneously aggregates the adversarial fooling loss, the Predictor’s prediction error, and demographic parity fairness loss.

\begin{equation} 
\mathcal{L}_{\text{Generator}} = \alpha  *\mathcal{L}_{\text{prediction}} + \beta  *\mathcal{L}_{\text{fooling}} +  \gamma  * \mathcal{L}_{\text{fairness}}
\end{equation}

% \begin{equation} 
% \alpha   + \beta   +  \gamma  = 1 
% \end{equation}
where \(G(x_i)\) is the Generator's output for the \(i\)th instance, \(D(G(x_i))\) is the Discriminator's prediction on this output and \(\alpha   + \beta   +  \gamma  = 1 \). This equation specifically aims to optimize the Generator's ability to produce data that the Discriminator classifies incorrectly, thereby enhancing privacy protection.

\begin{algorithm}[h]
\caption{Modified Adam Optimizer for Cost Function Coefficient Optimization}
\begin{algorithmic}[1]
\State Initialize hyperparameters $\alpha, \beta, \gamma$ with initial values.
\State Initialize Adam parameters $\alpha_m, \alpha_v, \beta_m, \beta_v, \gamma_m, \gamma_v$ to zero.
\State Initialize timestep $t$ to zero.
\State Define learning rate $\eta$, decay rates $\beta_1, \beta_2$, and a small constant $\epsilon$ for numerical stability.
\State Define loss contributions: $\mathcal{L}_{\text{prediction}}$, $\mathcal{L}_{\text{fairness}}$, and $\mathcal{L}_{\text{fooling}}$

\Procedure{Adam Update}{$param, grad, m, v, t$}
    \State $t \gets t + 1$
    \State $m \gets \beta_1 \cdot m + (1 - \beta_1) \cdot grad$
    \State $v \gets \beta_2 \cdot v + (1 - \beta_2) \cdot grad^2$
    \State $m_{hat} \gets \frac{m}{1 - \beta_1^t}$
    \State $v_{hat} \gets \frac{v}{1 - \beta_2^t}$
    \State $param \gets param - \eta \cdot \frac{m_{hat}}{\sqrt{v_{hat}} + \epsilon}$
    \State \Return $param, m, v, t$
\EndProcedure

\State Calculate total loss $total\_loss \gets \mathcal{L}_{\text{prediction}} +  \mathcal{L}_{\text{fooling}} +  \mathcal{L}_{\text{fairness}}$

\State Compute gradients based on loss proportions:
\State $grad_\alpha \gets -\left(\frac{\mathcal{L}_{\text{prediction}}}{total\_loss}\right)$
\State $grad_\beta \gets -\left(\frac{\mathcal{L}_{\text{fairness}}}{total\_loss}\right)$
\State $grad_\gamma \gets -\left(\frac{\mathcal{L}_{\text{fooling}}}{total\_loss}\right)$

\State $(\alpha, \alpha_m, \alpha_v, t) \gets \Call{AdamUpdate}{\alpha, grad_\alpha, \alpha_m, \alpha_v, t}$
\State $(\beta, \beta_m, \beta_v, t) \gets \Call{AdamUpdate}{\beta, grad_\beta, \beta_m, \beta_v, t}$
\State $(\gamma, \gamma_m, \gamma_v, t) \gets \Call{AdamUpdate}{\gamma, grad_\gamma, \gamma_m, \gamma_v, t}$

\State \textbf{return} $\alpha, \beta, \gamma, \alpha_m, \alpha_v, \beta_m, \beta_v, \gamma_m, \gamma_v, t $
\end{algorithmic}
\label{algo:adam_hyperparameter_update}
\end{algorithm}

% \subsection{Cost Function Optimization}
% Optimizing a cost function that balances fairness, privacy, and performance is a significant challenge in machine learning. Our initial approach involved an exhaustive search using the PFA framework to explore feasible balances, providing insights into the effects of the cost coefficients \(\alpha\), \(\beta\), and \(\gamma\). Based on these insights, we developed an optimization algorithm that adapts the Adam optimizer principles for efficient tuning of these coefficients. This method updates hyperparameters based on their associated losses, fine-tuning the cost function to achieve desired trade-offs in an efficient manner.
\subsection{Cost Function Optimization}
Optimizing a cost function that balanced fairness, privacy, and performance was a significant challenge in this multi-objective setting. The initial approach was based on an exhaustive search using the PFA framework to explore feasible trade-offs, which provided empirical insights into the effects of the cost coefficients \(\alpha\), \(\beta\), and \(\gamma\). Building on these observations, an optimization algorithm was then developed that adapted the Adam optimizer principles for efficient tuning of these coefficients. This procedure was intentionally used as a greedy, heuristic search over the weight space: rather than guaranteeing a single global optimum, it was designed to steer the model toward regions of the Pareto front where no objective could be improved without worsening at least one of the others, thereby yielding practically useful trade-offs in an efficient manner.

Our adaptive hyperparameter optimization updates hyperparameters based on the gradient of their associated losses, reflecting the proportion of each loss to the total loss. This allows for adaptive adjustments sensitive to the impact of each hyperparameter's performance, optimizing overall model performance. 
% We use an adapted version of the Adam optimizer to update \(\alpha\), \(\beta\), and \(\gamma\), crucial for controlling the trade-offs between fairness, privacy, and performance.
% To update \(\alpha\), \(\beta\), and \(\gamma\) effectively based on their associated losses, we employ a customized Adam optimization algorithm. 
Below is the pseudo code for the function update\_hyperparameters, outlining the procedure for adjusting these hyperparameters within our model.

\paragraph{Explanation of the Process}
The \texttt{update hyperparameters} function receives the current values of \(\alpha\), \(\beta\), and \(\gamma\), their related losses, and Adam optimizer settings. It calculates the combined total loss and computes each hyperparameter's gradient as the negative proportion of its associated loss to the total. For example, if the fairness loss contributes \(60\%\) of the total loss while the prediction and fooling losses contribute \(25\%\) and \(15\%\) respectively, the update step for \(\gamma\) will be roughly four times larger than those for \(\alpha\) and \(\beta\), thereby steering optimization toward reducing the fairness error that currently dominates overall performance.
%This ensures updates are sensitive to each component's impact on overall performance.

The \texttt{adam\_update} subroutine applies Adam optimization, using exponential moving averages of gradients to iteratively update hyperparameters, correcting biases in first and second moment estimates to adapt the learning rate, enhancing convergence. This method dynamically tunes hyperparameters, allowing adaptation to changes in data or model dynamics, thus improving training stability and performance.

\subsubsection{Implementation and Benefits}
This adaptive hyperparameter optimization proceeds through total loss calculation from all model loss components, gradient computation for each hyperparameter, and updates using the Adam rule to minimize overall loss.

\subsubsection{Advantages of the Adaptive Optimization Framework}
This adaptive optimization integrates hyperparameter adjustments directly with their impact on loss, enhancing specific model behaviors effectively and efficiently. The adaptation of the Adam optimizer ensures faster convergence using adaptive rates. 
% It offers scalability and flexibility, extending to additional hyperparameters or different loss functions suitable for diverse machine learning applications.
It also naturally mitigates overfitting by adjusting hyperparameters in response to comprehensive model performance feedback.
This strategy finely tunes each model performance aspect, enhancing fairness, privacy, and effectiveness without complicating the model structure.

\subsection{Model architecture}

The structure and the process of the main components of PFA architecture is illustrated in Figure \ref{fig:PFA}.

\textbf{Generator (G in Figure\ref{fig:PFA})}: 
    A seven‐layer feed-forward network (64 → 256 → 64) equipped with layer-wise normalization, dropout, a residual skip from the first hidden state, and a feature-wise attention gate stabilizes gradients, concentrates capacity on representational power on task-relevant features, and accelerates convergence while aggressively masking sensitive information.

\textbf{Predictor (P in Figure\ref{fig:PFA})}: 
    A five-layer MLP that expands to 128 neurons and then employs Leaky-ReLU activations to maintain gradient flow, yielding expressive yet tight representations that drive high classification accuracy with fast training dynamics.

\textbf{Discriminator (D in Figure\ref{fig:PFA})}: 
    A compact three-layer 32-64-32 architecture delivers a sharp adversarial signal capable of detecting residual sensitive cues, yet remains lightweight enough to keep the min--max optimization stable and efficient. 

Taken together, these three modules form an end-to-end adversarial architecture that are jointly optimized to balance accuracy, fairness, and privacy. Given an input $x_i$, the generator $G$ first produces a latent representation $z_i = G(x_i)$ in which task-relevant structure is preserved while information about the sensitive attribute is progressively suppressed. The predictor $P$ then consumes $z_i$ to produce the task prediction $\hat{y}_i = P(z_i)$ and is trained under an accuracy loss $\mathcal{L}_{\text{acc}}$ (e.g., cross-entropy between $y_i$ and $\hat{y}_i$), encouraging high utility on the primary task. In parallel, the discriminator $D$ attempts to infer the sensitive attribute via $\hat{s}_i = D(z_i)$ and is trained with a privacy loss $\mathcal{L}_{\text{priv}}$; gradients flowing from $\mathcal{L}_{\text{priv}}$ into $G$ are reversed so that $D$ is optimized to \emph{minimize} $\mathcal{L}_{\text{priv}}$, while $G$ is effectively optimized to \emph{maximize} it, driving $z_i$ to be uninformative about $s_i$. A fairness loss $\mathcal{L}_{\text{fair}}$ is additionally computed at the prediction level 
% (for example, a demographic-parity penalty based on group-wise differences in positive prediction rates), 
to align outcomes across protected and non-protected groups. 
% The overall training objective is given by  equation \ref{eq:total_loss} 
% \[
% \mathcal{L}_{\text{total}}
% = \alpha\, \mathcal{L}_{\text{acc}}
% + \beta\, \mathcal{L}_{\text{fair}}
% + \gamma\, \mathcal{L}_{\text{priv}},
% \]
% and was minimized with respect to $(G,P)$, while $D$ was updated only with respect to $\mathcal{L}_{\text{priv}}$. 
In this way, the latent space is simultaneously driven to be predictive of $y$, approximately invariant to $s$, and compliant with the chosen group-fairness criterion. At inference time, only $(G,P)$ are retained, so that the deployed model inherit the learned privacy–fairness–accuracy trade-off without additional adversarial overhead.

\color{black}

%%%%%%%%%%%%%%%%%%%%%%%%%%%%%%%%%%%%%%%%%%%%%%%%%%%%%%%%%%%%%%%%%%%%%%%%%%%%%%%%%%%%%%%%%%%%%

\section{Experimental Set up}
\subsection{Datasets}
We evaluate the performance of PFA using five datasets in order to cover a broad spectrum of fairness issues.
% To make the evaluation setting explicit,
The sensitive attributes, prediction targets, and fairness metrics used for each dataset were summarized in Table~\ref{tab:datasets}. For all datasets, a binary prediction task was defined, and at least one binary sensitive attribute was identified to operationalize group fairness. In our experiments, the fairness loss $\mathcal{L}_{\text{fair}}$ was instantiated primarily via a demographic-parity penalty on the positive class; for COMPAS, where equal-opportunity considerations are standard, an additional equal-opportunity gap was also reported.

\begin{table*}[t]
\centering
\caption{Summary of benchmark datasets, sensitive attributes, prediction tasks, and fairness metrics used in the experiments.}
\label{tab:datasets}
\begin{tabular}{|p{2.2cm}|p{2.7cm}|p{3.1cm}|p{2.2cm}|p{2.6cm}|}
\toprule
\textbf{Dataset} & \textbf{Prediction task ($y$)} & \textbf{Sensitive attribute} & \textbf{\#Instances / \#Features} & \textbf{Fairness metric(s)} \\
\midrule
Adult Income & Predict whether annual income $> \$50\mathrm{K}$ & Gender (Male vs. Female) & $\sim 48{,}000$ / 14 & Demographic-parity gap (positive prediction rate) \\
\addlinespace[0.3em]
Gender Pay Gap (GPG) & Predict high vs. low wage / salary band & Gender (Male vs. Female) & $344{,}287$ / 234 & Demographic-parity gap (positive prediction rate) \\
\addlinespace[0.3em]
COMPAS & Predict two-year recidivism risk & Gender (Male vs. Female) & $\sim 4{,}300$ / 27 & Demographic-parity gap;\newline Equal-opportunity gap (TPR difference) \\
\addlinespace[0.3em]
German Credit & Predict credit risk (Good vs. Bad) & Gender (Male vs. Female)  & $1{,}000$ / 20 & Demographic-parity gap (approval rate) \\
\addlinespace[0.3em]
MEPS (2015 Panel 19) & Predict high vs. low health-care expenditure / utilization &Gender (Male vs. Female) & $15{,}830$ / $1{,}815$ & Demographic-parity gap (positive outcome rate) \\
\bottomrule
\end{tabular}
\end{table*}
\color{black}

\paragraph{Baseline demographic bias.}
For each corpus, we quantify the \emph{inherent gender bias} %
\(b = \bigl|\,P(Y{=}1 \mid S=\text{category 1}) - P(Y{=}1 \mid S=\text{category 2})\,\bigr|\),%
\footnote{This is the empirical demographic-parity gap of the raw labels; %
a value of 0 indicates perfect balance, whereas larger values reflect wider historical %
inequity.
% that downstream models must mitigate.
} %
where \(S\) is the sensitive attribute \texttt{gender}.  
The resulting gaps are  
\textbf{0.1945} (Adult),  
\textbf{0.1455} (Gender Pay Gap),  
\textbf{0.1407} (COMPAS),  
\textbf{0.1189} (MEPS), and  
\textbf{0.0748} (German Credit).  
These figures highlight the diverse fairness challenges present before any learning takes place, which serve as reference points for the experimental results %
reported in Section 5.

\begin{table*}[!ht]
    \centering
    \caption{Comparative evaluation of DP\_G, DP\_L, PFairDP, FairDP and our method (PFA) across datasets. 
Bias is shown below each dataset name. Accuracy: higher is better. Fairness: lower is better. Privacy: closer to 50\% is better. \; Each entry reports \emph{mean $\pm$ standard deviation} computed over the five highest-accuracy runs (selected from 50 hyper-parameter configurations for PFA and PFairDP, 80 hyper-parameter configurations for DP\_G and DP\_L ) that met the filtering thresholds of demographic-parity $\le 5\%$ or closest to $5\%$ and from white box inference attack sensitive-attribute predictability $\le 65\%$ or closest to $65\%$ }%

    \begin{tabular}{|l|l|l|l|l|l|l|}
        \hline
        Dataset & Metric & DP\_G & DP\_L & PFairDP & FairDP & PFA (Ours) \\ \hline
        \multirow{3}{*}{\makecell[l]{Adult\\(Bias = 0.1945)}} 
            & Accuracy & 82.72\% $\pm$ 0.70\% & 82.90\% $\pm$ 1.45\% & 83.00\% $\pm$ 20.00\% & 78.70\% $\pm$ 1.81\% & \textbf{83.19\% $\pm$ 0.15\%} \\
            & Fairness & 0.078 $\pm$ 0.023 & 0.082 $\pm$ 0.037 & 0.060 $\pm$ 0.011 & 0.034 $\pm$ 0.002 & \textbf{0.016 $\pm$ 0.003} \\
            & Privacy  & 68.10\% $\pm$ 1.53\% & 68.96\% $\pm$ 3.83\% & 89.00\% $\pm$ 3.80\% & 70.01\% $\pm$ 2.10\% & \textbf{67.20\% $\pm$ 0.35\%} \\ \hline

        \multirow{3}{*}{\makecell[l]{COMPAS\\(Bias = 0.1407)}} 
            & Accuracy & 60.92\% $\pm$ 4.56\% & 60.95\% $\pm$ 6.76\% & \textbf{79.00\% $\pm$ 1.30\%} & 65.48\% $\pm$ 1.10\% & 67.07\% $\pm$ 1.13\% \\
            & Fairness & 0.069 $\pm$ 0.033 & 0.098 $\pm$ 0.036 & 0.120 $\pm$ 0.021 & 0.131 $\pm$ 0.023 & \textbf{0.036 $\pm$ 0.006} \\
            & Privacy  & 80.36\% $\pm$ 0.10\% & 80.26\% $\pm$ 0.09\% & 88.00\% $\pm$ 1.40\% & 68.23\% $\pm$ 3.80\% & \textbf{54.84\% $\pm$ 2.83\%} \\ \hline

        \multirow{3}{*}{\makecell[l]{Gender Pay Gap\\(Bias = 0.1455)}} 
            & Accuracy & 88.88\% $\pm$ 3.87\% & 88.82\% $\pm$ 6.50\% & 89.00\% $\pm$ 0.40\% & 90.07\% $\pm$ 0.51\% & \textbf{91.68\% $\pm$ 1.51\%} \\
            & Fairness & 0.112 $\pm$ 0.022 & 0.097 $\pm$ 0.046 & 0.090 $\pm$ 0.001 & 0.115 $\pm$ 0.032 & \textbf{0.014 $\pm$ 0.006} \\
            & Privacy  & 79.18\% $\pm$ 2.76\% & 77.27\% $\pm$ 6.87\% & 75.00\% $\pm$ 2.70\% & 75.43\% $\pm$ 4.72\% & \textbf{65.35\% $\pm$ 2.52\%} \\ \hline

        \multirow{3}{*}{\makecell[l]{German Credit\\(Bias = 0.0748)}} 
            & Accuracy & 81.00\% $\pm$ 7.56\% & 77.90\% $\pm$ 5.34\% & 53.57\% $\pm$ 4.64\% & 68.50\% $\pm$ 3.15\% & \textbf{81.78\% $\pm$ 12.38\%} \\
            & Fairness & 0.041 $\pm$ 0.045 & 0.055 $\pm$ 0.046 & 0.031 $\pm$ 0.020 & 0.029 $\pm$ 0.041 & \textbf{0.027 $\pm$ 0.034} \\
            & Privacy  & 81.70\% $\pm$ 6.07\% & 76.70\% $\pm$ 6.27\% & 73.63\% $\pm$ 11.58\% & 76.86\% $\pm$ 3.10\% & \textbf{62.71\% $\pm$ 7.59\%} \\ \hline

        \multirow{3}{*}{\makecell[l]{MEPS\\(Bias = 0.1189)}} 
            & Accuracy & 94.26\% $\pm$ 1.55\% & 95.04\% $\pm$ 2.31\% & 80.23\% $\pm$ 1.27\% & 72.52\% $\pm$ 5.79\% & \textbf{97.59\% $\pm$ 0.09\%} \\
            & Fairness & \textbf{0.018 $\pm$ 0.012} & 0.031 $\pm$ 0.031 & 0.018 $\pm$ 0.016 & 0.061 $\pm$ 0.025 & 0.126 $\pm$ 0.001 \\
            & Privacy  & 60.39\% $\pm$ 0.26\% & 60.51\% $\pm$ 0.53\% & 66.32\% $\pm$ 5.34\% & 62.75\% $\pm$ 2.94\% & \textbf{57.17\% $\pm$ 0.38\%} \\ \hline
    \end{tabular}
    % \caption{Comparative evaluation of DP\_G, DP\_L, PFairDP, and our method (PFA) across datasets. Bias is shown below each dataset name. Accuracy: higher is better. Fairness: lower is better. Privacy: closer to 50\% is better.}

    \label{results_table}
\end{table*}

\begin{table*}[htbp]
    \centering
     \caption{Fairness metrics for \textbf{PFA} across five datasets. 
    Columns report \emph{mean $\pm$ standard deviation}. Lower is better for all metrics. }
    \begin{tabular}{|l|c|c|c|c|}
        \hline
        \textbf{Dataset} & \textbf{Demographic Parity $\downarrow$} & \textbf{Equal Treatment $\downarrow$} & \textbf{Equal Opportunity $\downarrow$} & \textbf{Equalized Odds $\downarrow$} \\ \hline
        German Credit (GC) 
            & $0.027425 \pm 0.037554$ 
            & $0.025806 \pm 0.035337$ 
            & $0.025806 \pm 0.035337$ 
            & $0.071027 \pm 0.030993$ \\ \hline
        MEPS 
            & $0.126159 \pm 0.001442$ 
            & $0.009200 \pm 0.001300$ 
            & $0.011200 \pm 0.002300$ 
            & $0.021242 \pm 0.009841$ \\ \hline
        COMPAS 
            & $0.036 \pm 0.006$ 
            & $0.034 \pm 0.007$ 
            & $0.073 \pm 0.012$ 
            & $0.067 \pm 0.009$ \\ \hline
        Adult 
            & $0.016 \pm 0.003$ 
            & $0.153 \pm 0.160$ 
            & $0.110 \pm 0.159$ 
            & $0.057 \pm 0.038$ \\ \hline
        Gender Pay Gap (GPG) 
            & $0.013 \pm 0.006$ 
            & $0.334 \pm 0.047$ 
            & $0.334 \pm 0.047$ 
            & $0.078 \pm 0.025$ \\ \hline
    \end{tabular}
   
    \label{tab:pfa_fairness_breakdown}
\end{table*}

\subsection{Evaluation Metrics}

% \textbf{Class Prediction Performance using Accuracy:} Accuracy remains a standard metric for assessing the performance of classification models. It measures the proportion of true results (both true positives and negatives) among the total number of cases examined. For applications where correct predictions carry significant consequences, such as healthcare diagnostics, high accuracy ensures that the models perform their intended tasks effectively

\textbf{Class prediction performance using accuracy:} Accuracy was used as the primary measure of class prediction performance. In all experiments, accuracy was computed from the outputs of the main task classifier, i.e., from predictions $\hat{y}_i = P(G(x_i))$ on the held-out test set, where $G$ denoted the generator and $P$ denoted the predictor. 
% Formally, if $\mathcal{D}_{\text{test}}$ denoted the test set,
% \begin{equation}
% \label{eq:accuracy}
% \text{Acc} = \frac{1}{|\mathcal{D}_{\text{test}}|} \sum_{(x_i,y_i) \in \mathcal{D}_{\text{test}}} \mathbf{1}\big[\hat{y}_i = y_i\big].
% \end{equation}
This metric thus reflects how well the composed model $P \circ G$ performed the intended classification task, while the discriminator $D$ only influenced the learning of $G$ via the privacy loss and was not used directly for reporting class prediction accuracy.
\color{black}

\textbf{Fairness Evaluation through Demographic Parity:} Demographic parity calculates the probability of a positive outcome across groups that are defined by the sensitive attribute. This metric is crucial because it focuses on non-discrimination and mitigating historical biases. 

\textbf{Proposition 1 (Demographic parity blocks transfer of historical base rates).}
Let $S\!\in\!\{0,1\}$ be sensitive and suppose historical labels are biased so that $P(Y{=}1\mid S{=}0)\neq P(Y{=}1\mid S{=}1)$. Any predictor $\hat Y$ satisfying demographic parity (DP),
\begin{equation}
\label{eq:demographic_parity}
P(\hat{Y} = 1 \mid S = 0) = P(\hat{Y} = 1 \mid S = 1),
\end{equation}

% equalizes selection rates and therefore \emph{cannot} reproduce those historical disparities in outcomes. \emph{Proof.} Immediate by definition, since DP makes $\hat Y\!\perp\!\!\!\perp S$ at deployment, independent of $P(Y\mid S)$. \qed
equalizes selection rates and therefore \emph{cannot} reproduce those historical disparities in outcomes.
\emph{Proof.} Demographic parity (DP) requires the predictor to satisfy
\[
\hat Y \;\perp\!\!\!\perp\; S,
\]
i.e., the deployed decision $\hat Y$ is statistically independent of the sensitive attribute $S$.
Equivalently, for any two groups $s,s'$ in the support of $S$ and any decision value $\hat y$,
\[
\Pr(\hat Y=\hat y \mid S=s) \;=\; \Pr(\hat Y=\hat y \mid S=s').
\]
In the common binary-decision setting ($\hat Y\in\{0,1\}$), this reduces to equality of selection rates:
\[
\Pr(\hat Y=1 \mid S=s) \;=\; \Pr(\hat Y=1 \mid S=s').
\]
Hence, at deployment, the model cannot produce different acceptance/selection rates across groups due solely
to $S$, regardless of any historical disparity in the base rates $P(Y\mid S)$.

In contrast to fairness metrics that rely on ground truth labels, thus deeming a model that perfectly reproduces historically biased outcomes ``fair" and reinforcing those biases\cite{denis2024fairness}, demographic parity assesses equity without referencing the true label distribution. For instance, if a system outputs $\hat Y\!=\!Y$, then the $\mathrm{TPR}\!=\!1$ and $\mathrm{FPR}\!=\!0$ for all groups perfect equalized odds~\cite{hardt2016equality}, yet $P(\hat Y{=}1\mid S)=P(Y{=}1\mid S)$ \emph{replicates} historical disparities. DP forbids this by construction. This label-agnostic approach blocks the direct transfer of historical discrimination into new systems by requiring equal positive prediction rates across sensitive groups, thereby actively discouraging the persistence of entrenched inequities.

% Why Demographic Parity is mathematically justified.

\textbf{Proposition 2 (Robustness to label corruption).}
Let $\tilde Y$ be produced from $Y^\star$ via group-dependent noise $P(\tilde Y\mid Y^\star,S)$. DP of $\hat Y$ depends only on $(\hat Y,S)$ and is invariant to such corruption, whereas label-conditioned metrics Equalized Odds/Equal Opportunity (EO/EOp) can be arbitrarily distorted~\cite{chouldechova2017fair,kleinberg_inherent_2017}. 
\emph{Implication:} when labels encode historical disadvantage, DP provides a label-agnostic safeguard; trade-offs with calibration/utility follow from known impossibility results~\cite{chouldechova2017fair,kleinberg_inherent_2017}.

% \textbf{Privacy Assessment through Predictability - from white box inference attack:} The predictability of the sensitive variable measures how well a model can infer sensitive attributes from its outputs. Low predictability implies the outputs do not reveal much about the underlying sensitive attributes, thus better preserving privacy. 

\textbf{Privacy assessment through predictability (white-box inference attack):} The predictability of the sensitive variable is used to quantify privacy, by measuring how well an attacker could infer sensitive attributes from the model outputs. In this setting, a white-box inference attacker is instantiated using a CatBoost classifier.
% whose hyperparameters are optimized on a validation split. 
The attacker is trained on the outputs of the PFA model (predicted scores derived from $P(G(x_i))$) as input features and the true sensitive labels $s_i$ as targets. Formally, a separate attack model $A$ is fitted to approximate $s_i \approx A(P(G(x_i)))$, and its classification performance (e.g., in terms of prediction accuracy or related metrics) is interpreted as a measure of information leakage. Lower attack performance indicates that the it is harder to interpret the sensitive attribute, and thus that stronger privacy is achieved.
% by the learned representation.
\color{black}

\subsection{Baselines}
Recent studies propose alternatives such as counterfactual adversarial embeddings for recommender systems \cite{10.1145/3643670}, Wasserstein Differential Privacy aimed solely at tightening leakage bounds \cite{DBLP:conf/aaai/YangQZ24}, and FairDP-SGD/FairPATE, which explore privacy–fairness trade-offs in federated settings \cite{yaghini2023learning}.  
We do not adopt these as baselines because (i) they target specialized domains or decentralized training regimes distinct from our tabular, centralized classification task, (ii) reproducible implementations or hyper-parameter grids are unavailable for a controlled comparison, and (iii) they target at most two objectives either accuracy with privacy or accuracy with fairness instead of jointly optimizing all three: privacy, fairness, and accuracy.
%they optimize at most two , accuracy with privacy or fairness  rather than all three objectives of privacy, fairness, and accuracy.  
Accordingly, we benchmark PFA against PFairDP, plus canonical Gaussian and Laplacian noise mechanisms, which together span the principal design space relevant to our setting.

\textbf{PFairDP~\cite{ficiu2023automated}:} This method integrates differential privacy directly within the training of the model, aiming to balance privacy and fairness without significantly degrading the predictive performance. 

\textbf{FairDP} study how differential privacy (DPriv) impacts group fairness in classification. They prove fairness metrics are Lipschitz in model parameters and bound the gap between private and non-private learners under output perturbation and DP-SGD, shrinking as $\tilde{O}(\sqrt{p/n})$. \cite{mangold2023differential}.

\textbf{Gaussian Noise Addition (DP-G):} A standard approach in differential privacy involves incorporating Gaussian noise into the dataset to obscure sensitive attributes, thereby protecting individual data points. We test the privacy, fairness and utility for epsilon values systematically varying from \(10^{-5}\) to \(10\), following a sequence: \(10^{-5}\), \(\sqrt{10} \cdot 10^{-5}\), \(10^{-4}\), \(\sqrt{10} \cdot 10^{-4}\), \ldots, \(\sqrt{10}\), \(10\). This approach ensures a comprehensive exploration of privacy settings to assess the trade-offs between data utility and privacy. 

\textbf{Laplacian Noise Addition (DP-L):} Similar to Gaussian noise, Laplacian noise is integrated to ensure privacy. The Laplacian distribution, characterized by its sharper exponential decay, is employed as it might offer enhanced effectiveness in scenarios where privacy requirements are stringent.

\section{Results \& Discussion}
\begin{figure*}[t]
\includegraphics[width=17cm]{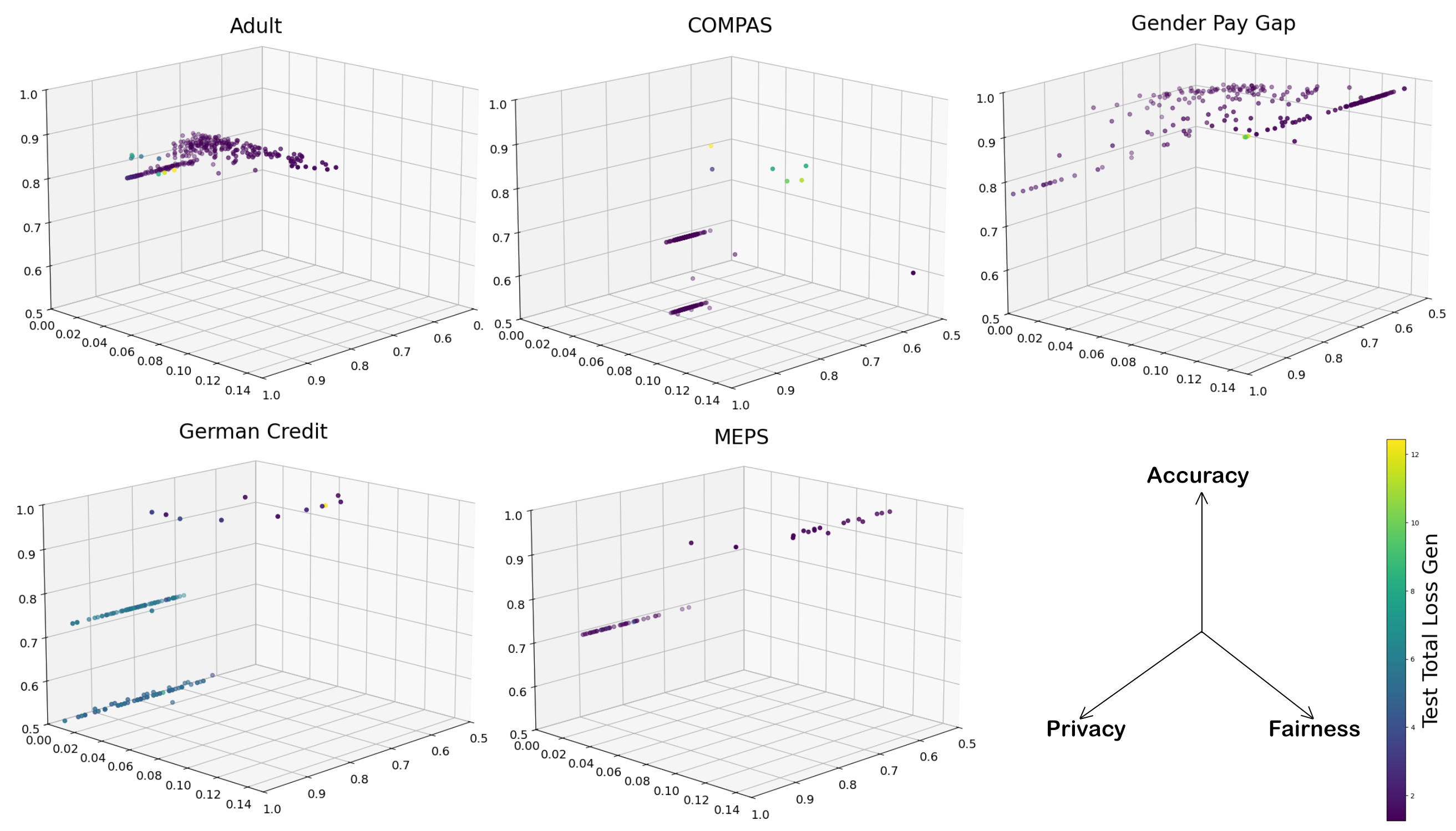}
  \caption{Three-dimensional scatter plots visualizing performance of the models for the 5 datasets. The axes convey the three evaluation criteria \textit{fairness} (x, lower is better), \textit{privacy} (y, values close to 50\% are better) and \textit{accuracy} (z, higher is better). Each marker denotes one hyper-parameter configuration; its color encodes the generator’s test-set loss, with darker shades indicating a lower loss and therefore a milder accuracy–fairness–privacy trade-off. Clusters of dark points near the ideal corners of the cube signal configurations that approach all three objectives simultaneously.}
  \label{fig:Results}
\end{figure*}

\subsection{Comparative Analysis of PFA vs.\ Baselines}

Table~\ref{results_table} contrasts our PFA framework with three baselines across \emph{five} public benchmarks that range from 
\textit{Adult} \& \textit{Gender Pay Gap} income disparity, \textit{COMPAS} criminal risk file,  \textit{German Credit} portfolio, and \textit{MEPS19} healthcare panel. 

\textbf{Adult} %(14 features, bias 0.1945).
On the Adult dataset PFA attains the highest predictive accuracy (83.2\%), the smallest demographic-parity gap (1.6\%), and the lowest from white box inference attack sensitive-attribute predictability (67.2\%, i.e., only 17.2\% above random), thereby surpassing all three baselines in the joint privacy–fairness–accuracy space.

\textbf{COMPAS} %(27 features, bias 0.1407).
The significantly skewed label distribution ( 80.9\% male labels) and the limited size of COMPAS hamper every method: PFairDP records the top accuracy (79\%) but with a 12\% parity gap and weak privacy. PFA accepts a \(\approx\)12\% accuracy reduction to cut unfairness to 3.6\% and nearly halve sensitive-attribute predictability, exemplifying an explicit fairness accuracy compromise.

\textbf{Gender Pay Gap} %(234 features, bias 0.1455).
Despite the dataset’s high dimensionality and class imbalance, PFA improves accuracy by almost three percentage points relative to the closest competitor while reducing unfairness by a factor of seven, again delivering the strongest overall trade-off.

\textbf{German Credit}% (20 features, bias 0.0748).
In the small-sample German Credit benchmark, PFA simultaneously secures the best accuracy and the lowest parity gap, all while preserving the tightest privacy budget; DP\_L and PFairDP, by contrast, incur substantial performance degradation.

\textbf{MEPS}  
Although PFA achieves near-perfect accuracy (97.6\%) and the most privacy-protective outcome (57.2\%), its 12.6\% demographic-parity gap remains above that of the label-dependent DP\_G baseline 1.8\%.

\noindent\textit{Observation on PFairDP robustness.}
PFairDP attains the highest accuracy on \textsc{COMPAS}, but it does not lead on the other benchmarks considered here. The divergence is informative: \textsc{COMPAS} is relatively small, label-imbalanced, and structurally distinct from the higher-dimensional \textsc{GPG}/\textsc{MEPS} and the low-sample \textsc{German Credit} settings. PFairDP’s favorable behavior on \textsc{COMPAS} thus appears to depend on dataset-specific characteristics rather than transferring across regimes. Moreover, its fairness and privacy outcomes on non-\textsc{COMPAS} corpora are not competitive with the best-performing methods. Taken together, these observations indicate limited robustness of PFairDP when evaluated under heterogeneous data distributions, feature scales, and class priors, and motivate approaches that better preserve performance stability under such shifts.

\noindent\textit{FairDP behavior across datasets.}
FairDP exhibits comparatively larger demographic-parity gaps on most datasets, with two notable exceptions: \textsc{Adult} and \textsc{German Credit}, where its fairness is competitive. The method’s performance degrades on \textsc{GPG} and \textsc{MEPS}, which combine higher dimensionality with complex covariate structure, and on \textsc{COMPAS}, where label imbalance is pronounced. This pattern suggests sensitivity to data distributional properties and limited invariance across domains. While FairDP can be effective in specific regimes, the cross-dataset evidence indicates that its fairness advantages are not consistent. In the context of system selection for deployment, such variability underscores the importance of evaluating fairness methods under multiple corpora to assess robustness rather than relying on single-benchmark gains.

In summary, PFA offers the most favorable privacy guarantee on every benchmark, leads the accuracy ranking on four of five datasets, and delivers the lowest unfairness on three, thereby demonstrating robust generalization across heterogeneous data regimes.

In particular, when jointly assessing (i) demographic parity, where smaller gaps are preferable, (ii) white-box inference attack sensitive-attribute predictability, where values closer to $50\%$ indicate reduced leakage, and (iii) task accuracy/AUC, higher being better, the experiments here place PFA near the preferred region of this three-objective space. The baselines tend to optimize at most two of these criteria simultaneously, whereas PFA maintains competitive utility while lowering both demographic disparity and sensitive-attribute predictability. This joint evaluation, rather than any single metric, supports the conclusion that PFA provides a more stable balance across privacy, fairness, and accuracy under the tested data regimes.

\subsection{Privacy, Fairness, and Accuracy Trade-offs}

The analysis of our Privacy-Fairness-Accuracy (PFA) model's performance, depicted in Figure \ref{fig:Results}, demonstrates how effectively it navigates the trade-offs between accuracy, privacy, and fairness. For example, in the Gender Pay Gap panel, a marker located near the inner-top corner of the cube corresponds to a configuration with 92.2\% accuracy, a demographic-parity gap of 0.008, and the inference attack predictability of 64.8\%.  Its position high on the accuracy axis, low on the fairness axis, and close to the ideal 50\% privacy plane illustrates the type of balanced operating point the framework seeks to uncover. The model aims for high accuracy in class predictions, minimal fairness loss to ensure equitable treatment across demographic groups, and reduced predictability of sensitive attributes to enhance privacy.

On the \textbf{Adult dataset}, PFA exhibits a notably well-balanced performance, simultaneously sustaining competitive accuracy, stringent privacy, and a low demographic-parity gap. Such equilibrium renders the model particularly suitable for settings in which equitable outcomes must be secured without materially diminishing predictive utility or privacy safeguards. Although DP-G or PFairDP occasionally achieve marginal gains on individual metrics, PFA delivers the best aggregate outcome, offering the most balanced trade-off among accuracy, privacy, and fairness across all evaluations.

On the \textbf{Gender Pay Gap dataset}, PFA likewise maintains strong harmony: accuracy consistently exceeds 91 \%, the demographic-parity gap contracts to 0.008, and white-box inference attack sensitive-attribute predictability  stabilizes near the 65 \% target. This shows that substantial fairness gains can be realized with only a negligible loss in utility and no privacy penalty an outcome unattained by competing baselines.

On the \textbf{COMPAS dataset}, where 80.9 \% of records are male, PFA deliberately forgoes about 12 percentage points of accuracy to reduce the parity gap to 0.036 and cut predictability to 55 \%. No baseline offers a comparably equitable and privacy-respectful compromise under such extreme label imbalance.

On the \textbf{German Credit dataset}, despite the small sample of 1 000 cases, PFA achieves a ridge around 82 \% accuracy, a 0.027 parity gap, and 63 \% predictability, illustrating its capacity to balance all three objectives even in data-sparse conditions.
% where the baselines deteriorate.

On the \textbf{MEPS dataset}, encompassing 1 815 features, PFA sustains near-perfect 97 \% accuracy while holding predictability at 57 \% and the parity gap at 0.126. Attempts to push fairness further erode accuracy sharply, confirming that PFA locates the optimal trade-off point that baselines fail to achieve.

Across all panels, the darkest region consistently resides in the cube’s interior high accuracy, Demographic parity well below the raw-data bias, and inference attack sensitive predictability driven toward random-guess levels. This pattern confirms that PFA does not merely perform well in isolated metrics but instead identifies configurations that balance all three objectives in a principled, data-dependent manner.
Overall, when evaluating the balance of privacy, fairness, and accuracy, the PFA model distinguishes itself as a comprehensive solution. Although PFairDP also targets these three objectives, PFA consistently demonstrates superior adaptability across various data environments and operational requirements, often outperforming standard approaches that might prioritize one metric over others. 

\begin{figure*}[!htbp]
  \includegraphics[width=16.5cm]{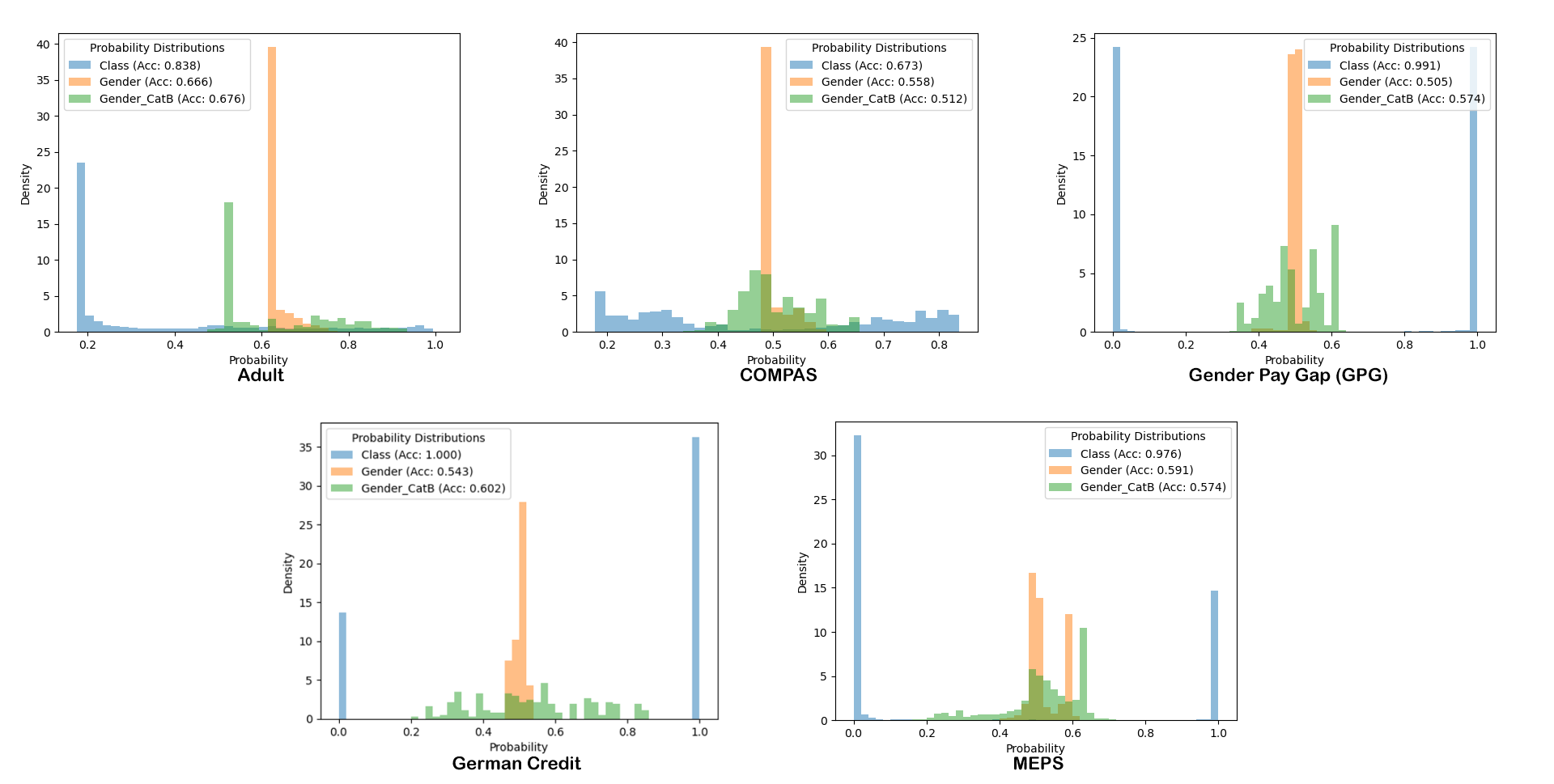}
  \caption{The prediction confidence distributions for the class and sensitive attributes across five datasets using the test-set.
  The blue distribution represents the class prediction probability distribution, which ideally exhibits a bimodal distribution with peaks at the extremes of the probability scale.
  The orange distribution corresponds to the discriminator's prediction probability distribution, whilst the green distribution captures the sensitive attribute prediction probability.
  The orange and green distributions should ideally follow a tight normal distribution centered around 0.5.
  \emph{Reading example – GPG panel:} the blue histogram peaks near 0 and 1, whereas both orange and green distributions form a narrow mound around 0.5, signifying that the model achieves confident class predictions while revealing minimal gender information.}
  \label{fig:privacy_distribution_exp}
\end{figure*}

\subsection{Prediction Confidence}
Simply measuring privacy by a model’s overall success for predicting a sensitive binary attribute can be misleading. Achieving 50\% predictability may look like random guessing, yet the true privacy risk lies in how confidently the model makes predictions for certain subgroups. If prediction confidences are unevenly distributed, some individuals or groups could be more easily identified, undermining fairness. As illustrated in Figure \ref{fig:privacy_distribution_exp}, the prediction distributions for the Adult and GPG datasets indicate central tendencies near a mean of 0.5, with tightly clustered distributions for the sensitive attribute, gender. Remarkably, when utilizing the optimized CatBoost algorithm, the predictability distributions become notably narrower, underscoring CatBoost's utility as a robust method for assessing predictability. Moreover, the distributions for class predictions across these datasets display extensive variability and high standard deviations, with pronounced peak values at the distribution's extremes (0 and 1), affirming that essential class prediction information is retained effectively in the synthetic datasets.

\subsection{Hyper-parameter Tuning Runtime}

\begin{figure}
  \centering
  \includegraphics[width=0.9\linewidth]{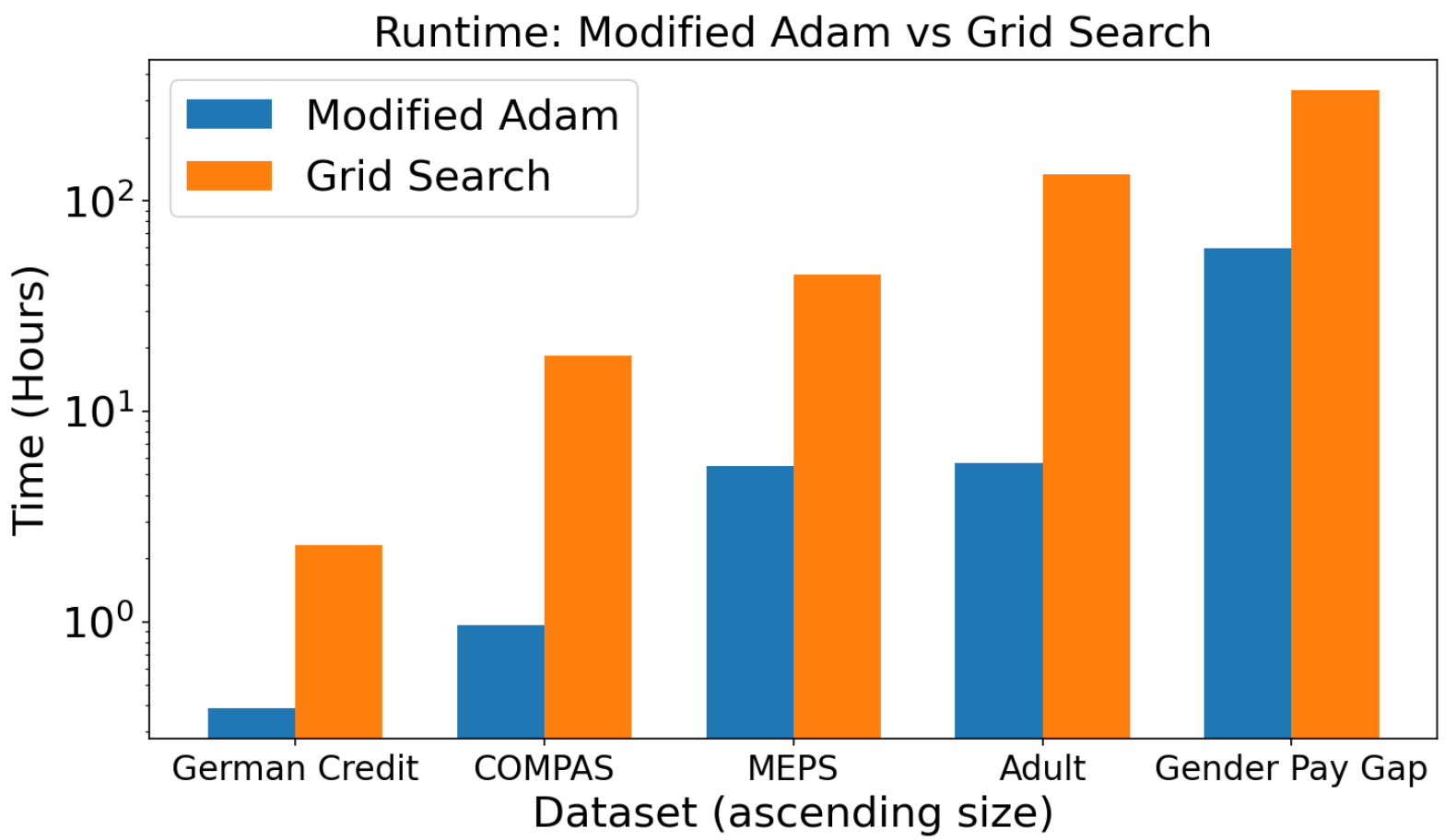}
  \caption{Wall-clock time (log scale) to find a configuration that meets the
  target thresholds of demographic-parity $\le 5\%$ and
  white-box inference attack sensitive-attribute predictability $\le 65\%$.  
  }
  \label{fig:runtime}
\end{figure}

Figure \ref{fig:runtime} benchmarks the end-to-end wall-clock time required to \emph{discover} a qualifying set of cost-function weights.
% on identical hardware. 
A grid search, often the default strategy, trains a fresh model for \emph{every} point on a dense $21^3$ lattice in $(\alpha,\beta,\gamma)$ space; even on the 1,000 row German Credit corpus, this entails hours of computation, and for GPG it exceeds ten days.  Our Modified Adam scheduler instead performs 50 sequential updates, shifting probability mass toward promising regions while training a single model per step.  The result is a $16\times$ speed-up on German Credit and a $>50\times$ reduction on GPG.  

% Place this new subsection in Section V (Results \& Discussion)
% immediately AFTER Subsection B (Privacy, Fairness, and Accuracy Trade-offs)
% and BEFORE the current Subsection C (Prediction Confidence).
% If you keep alphabetical labels, rename the subsequent subsections accordingly.

\subsection{
% Interpreting Fairness Metrics 
% in Table~\ref{tab:pfa_fairness_breakdown} and 
Retargeting the PFA Fairness Objective}

We formalize the four group metrics reported in Table~\ref{tab:pfa_fairness_breakdown}.  
Let $S\in\{0,1\}$ denote the sensitive attribute, $Y\in\{0,1\}$ the ground-truth label, and $\hat{Y}=g(f(x))\in\{0,1\}$ the model prediction.  
\emph{Demographic Parity (DP)} measures selection-rate parity:

\begin{equation}
\label{eq:dp_gap}
\mathrm{DP} \;=\; \big|\Pr(\hat{Y}=1\mid S=0)-\Pr(\hat{Y}=1\mid S=1)\big|.
\end{equation}

\emph{Equal Treatment (ET)} is defined here as the error-rate gap:

\begin{equation}
\label{eq:et_gap}
\mathrm{ET} \;=\; \big|\Pr(\hat{Y}\neq Y\mid S=0)-\Pr(\hat{Y}\neq Y\mid S=1)\big|.
\end{equation}

\emph{Equal Opportunity (EOp)} compares true–positive rates:

\begin{equation}
\label{eq:equal_opportunity}
\mathrm{Equal\ Opportunity} \;=\; \big|\mathrm{TPR}_0-\mathrm{TPR}_1\big|
\end{equation}

\begin{equation}
\label{eq:equal_opportunity_sub}
\mathrm{TPR}_s=\Pr(\hat{Y}=1\mid Y=1,S=s).
\end{equation}

\emph{Equalized Odds (EOdds)} jointly aligns true–positive and false–positive rates:
\begin{equation}
\label{eq:equal_odds}
\mathrm{Equal\ Odds} \;=\; \frac{1}{2}\Big(\big|\mathrm{TPR}_0-
\mathrm{TPR}_1\big|+\big|\mathrm{FPR}_0-\mathrm{FPR}_1\big|\Big),
\end{equation}

\begin{equation}
\label{eq:equal_odds_sub}
\mathrm{FPR}_s=\Pr(\hat{Y}=1\mid Y=0,S=s).
\end{equation}

Our PFA objective instantiates the fairness term $L_{\text{fairness}}$ using a differentiable proxy for DP (Sec.~III-B), so we expect low DP gaps when optimization pressure is effective. This aligns with the results for \textsc{Adult} (0.016), \textsc{GPG} (0.013), \textsc{German Credit} (0.027), and \textsc{COMPAS} (0.036), where selection rates are close across groups. In contrast, on \textsc{MEPS} the DP gap remains higher (0.126) while the label conditioned metrics are small (\textsc{Equal Opportunity}$\approx 0.011$, \textsc{Equalized Odds}$\approx 0.021$). This pattern is consistent with known incompatibilities among criteria under differing base rates and calibration constraints~\cite{chouldechova2017fair,kleinberg_inherent_2017,hardt2016equality}: when groups have distinct prevalences, pushing one notion (e.g., selection parity) can conflict with perfect alignment of error components.

Because $L_{\text{fairness}}$ is modular, PFA can be \emph{retargeted} to other metrics by replacing the DP surrogate with differentiable penalties for Equal Treatment, Equal Opportunity, or Equalized Odds. Concretely, for Equal Opportunity one may penalize the absolute difference of groupwise mean scores on the $Y{=}1$ subset; for Equalized Odds, combine Equal Opportunity with an analogous penalty on the $Y{=}0$ subset; for Equal Treatment, penalize the groupwise error-rate gap via smooth surrogates to 0–1 loss. The rest of the architecture and training loop remain unchanged. Hence, if the application prioritizes equal opportunity (e.g., recall parity in high-stakes detection), the fairness term can be swapped accordingly without redesigning the generator–predictor–discriminator triad.

A practical consideration is how the choice of metric interacts with utility. Equal Opportunity and Equalized Odds are \emph{label-conditioned}: they explicitly reward correct classification on positives (Equal Opportunity) and balance both positives and negatives (Equalized Odds). When labels are reliable, optimizing these criteria can improve task performance relative to strict DP, because the fairness signal is aligned with the decision boundary that separates $Y{=}1$ from $Y{=}0$. This alignment is visible in Table~\ref{tab:pfa_fairness_breakdown}: on \textsc{MEPS}, despite a larger DP gap, the low Equal Opportunity/Equalized Odds  indicate that predictions are well-calibrated across groups conditional on $Y$, which typically preserves accuracy. By contrast, DP is label-agnostic: it promotes equal selection rates irrespective of base rates or label quality. This property helps block the direct transfer of historical disparities when labels are biased or structurally confounded, but it can reduce accuracy in settings where group prevalences genuinely differ and labels are trustworthy.

These trade-offs counsel an application-specific choice. If the central risk is the reproduction of historical disadvantage (e.g., hiring with skewed past outcomes), DP offers a label-agnostic safeguard. If the objective is parity of error components in well-specified supervised tasks (e.g., screening with vetted labels), Equal Opportunity or Equalized Odds may be preferable and can be implemented in PFA by changing only the fairness penalty. Finally, Equal Treatment provides a compact view of overall error-rate parity and can be useful for auditing deployment-time behavior.

In summary, Table~\ref{tab:pfa_fairness_breakdown} reflects the fact that PFA has been tuned for DP, yielding small selection-rate gaps on four datasets and competitive label-conditioned behavior on others. Because the fairness term is pluggable, the same framework can target Equal Treatment, Equal Opportunity, or Equalized Odds when those definitions are better aligned with domain requirements potentially improving utility when labels are reliable while recognizing the impossibility results that preclude simultaneous optimization of all criteria when base rates differ.

%%%%%%%%%%%%%%%%%%%%%%%%%%%%%%%%%%%%%%%%%%%%%%%%%%%%%%%%%%%%%%%%%%%%%%%%%%%%%%%%%%%%%%%%%%%%%

\subsection{Ablation Study}

\begin{table}[htbp]
\centering
\caption{Performance results under different weight settings of $\alpha$, $\beta$, and $\gamma$ in the loss function.Accuracy: higher is better. Fairness: lower is better. Privacy: closer to 50\% is better.}
\renewcommand{\arraystretch}{1}  % optional row padding
\begin{tabular}{|l|l|l|l|l|l|l|}
\hline
Dataset & $\alpha$ & $\beta$ & $\gamma$ &  Accuracy & Fairness & Privacy \\ \hline
Adult            & 0   & 0.5   & 0.5   & \textbf{0.763} & 0.000 & 0.964 \\ 
                 & 0.5   & 0   & 0.5   & 0.909 & \textbf{0.088} & 0.677 \\ 
                 & 0.5   & 0.5   & 0   & 0.738 & 0.000 & \textbf{1.000} \\ \hline
GPG              & 0   & 0.5   & 0.5   & \textbf{0.770} & 0.000 & 0.734 \\ 
                 & 0.5   & 0   & 0.5   & 0.843 & \textbf{0.005} & 0.614 \\ 
                 & 0.5   & 0.5   & 0   & 0.771 & 0.000 & \textbf{0.997} \\ \hline
COMPAS           & 0   & 0.5   & 0.5   & \textbf{0.488} & 0.000 & 0.901 \\ 
                 & 0.5   & 0   & 0.5   & 1.000 & \textbf{0.104} & 0.841 \\ 
                 & 0.5   & 0.5   & 0   & 0.998 & 0.037 & \textbf{0.998} \\ \hline
MEPS             & 0 & 0.5 & 0.5 & \textbf{0.296} & 0.000 & 0.905 \\ 
                 & 0.5 & 0 & 0.5 & 0.999 & \textbf{0.119} & 0.913 \\ 
                 & 0.5 & 0.5 & 0 & 0.999 & 0.118 & \textbf{0.999} \\ \hline
German Credit    & 0 & 0.5 & 0.5 & \textbf{0.323} & 0.000 & 0.995 \\ 
                 & 0.5 & 0 & 0.5 & 0.995 & \textbf{0.133} & 0.973 \\ 
                 & 0.5 & 0.5 & 0 & 1.000 & 0.143 & \textbf{1.000} \\ \hline
\end{tabular}

\label{ab_study}
\end{table}

Table~\ref{ab_study} demonstrates that tuning the loss–function weights
\( \alpha \), \( \beta \), and \( \gamma \) provides explicit control over
\emph{prediction accuracy}, \emph{fairness}, and \emph{privacy}, respectively.
Across all datasets, the rows with \( \alpha = 0 \) (first row per dataset) shows
the lowest accuracies (e.g., \(0.488\) on \textsc{COMPAS} and \(0.296\) on
\textsc{MEPS}), whereas increasing \( \alpha \) to \(0.5\) consistently raises
class–prediction performance (e.g., accuracy on \textsc{COMPAS} increased from
\(0.488\) to \(1.000\) when \((\alpha,\beta,\gamma)\) changed from
\((0,0.5,0.5)\) to \((0.5,0,0.5)\), and from \(0.296\) to \(0.999\) on
\textsc{MEPS}). This  suggests that larger \( \alpha \) values steered the
model toward utility, even when fairness or privacy terms are present.

The effect of the fairness weight \( \beta \) was most clearly seen when comparing
the second and third rows within each dataset. Setting \( \beta = 0 \) increases
the demographic–parity gap, whereas reintroducing \( \beta = 0.5 \) reduces it.
For example, on \textsc{Adult}, the fairness error rose from \(0.000\) to
\(0.088\) when moving from \((0,0.5,0.5)\) to \((0.5,0,0.5)\), and then dropped
back to \(0.000\) under \((0.5,0.5,0)\). A similar behaviour was observed on
\textsc{COMPAS}, where the fairness gap increased to \(0.104\) when
\(\beta = 0\) and decreased to \(0.037\) once \(\beta\) was restored to \(0.5\).
These trends confirmed that the fairness branch effectively regularized the
predictor toward group parity.

% The privacy weight \( \gamma \) was ablated explicitly in this study to assess
% its impact on information leakage.
For every dataset, the configuration with
\(\gamma = 0\) (third row) removes the privacy term from the objective and
produces the largest predictability scores, corresponding to the strongest leakage
(e.g., \textsc{Adult}: \(1.000\); \textsc{GPG}: \(0.997\); \textsc{MEPS}:
\(0.999\)), meaning that the white–box CatBoost attacker could almost perfectly
predict the sensitive attribute. When the privacy term is reintroduced with
\(\gamma = 0.5\) (first and second rows), the attack accuracy moved closer to the
random baseline of \(50\%\) for a binary sensitive attribute (e.g., on
\textsc{Adult}, privacy decreases from \(1.000\) to \(0.964\) and \(0.677\)),
showing that \(\gamma > 0\) systematically reduced the attacker's ability to
recover the sensitive variable from the model outputs.

% Taken together, these three coefficients acted as complementary “compartments”
% in the objective: \( \alpha \) steered utility, \( \beta \) enforced fairness,
% and \( \gamma \) safeguarded privacy. The concrete trade–offs visible in
% Table~\ref{ab_study}-for instance, the perfect accuracy but non–negligible
% fairness gap on \textsc{COMPAS} under \((\alpha,\beta,\gamma) = (0.5,0,0.5)\),
% or the stronger privacy but reduced accuracy on \textsc{Adult} under
% \((0,0.5,0.5)\)-illustrated why all three components needed to be balanced.
% Adjusting the triplet \((\alpha, \beta, \gamma)\) therefore provided a
% principled lever for navigating the accuracy–fairness–privacy frontier within
% this architecture.

\color{black}

%%%%%%%%%%%%%%%%%%%%%%%%%%%%%%%%%%%%%%%%%%%%%%%%%%%%%%%%%%%%%%%%%%%%%%%%%%%%%%%%%%%%%%%%%%%%%

\subsection{Discussion}

The empirical evaluations presented in this study underscore the critical interplay between privacy, fairness, and accuracy within the context of the developed PFA framework. 

 PFA demonstrates robust capabilities in managing privacy and fairness simultaneously across diverse datasets such as Adult and GPG. This balance is facilitated by the innovative use of adversarial learning mechanisms, in which the generator and the discriminator collaboratively obscure sensitive attributes. Such dynamic manipulation ensures that sensitive information remains protected, addressing prevalent concerns around data privacy in AI applications.

However, the study also highlights the PFA model's challenges with severely imbalanced datasets, notably the COMPAS dataset, where the model exhibited biases towards the majority class. This limitation suggests areas for future improvement, particularly in enhancing the model's ability to handle data disparities effectively. The successful adjustment observed in the COMPAS (Balanced) scenario against the original unbalanced distribution (Figure~\ref{fig:privacy_distribution_exp}, right) suggests that pre-processing techniques aimed at data balance enhancement could be integral to improving fairness outcomes.

Moreover, the comparative analysis between traditional differential privacy methods, such as Gaussian and Laplacian noise applications, and the PFA approach highlights the inherent trade-offs involved. While traditional methods provide robust privacy by masking individual data points, they often do so at the expense of accuracy and fairness. The PFA model, conversely, offers a nuanced approach that more precisely controls these trade-offs, thereby preserving data utility without compromising privacy.

%\noindent\textit{Synthesis from the ablation study.}
The ablation analysis indicates that each model component materially contributes to meeting the simultaneous privacy–fairness–utility targets. Removing or down-weighting the generator’s adversarial objective increases white-box inference attack sensitive-attribute predictability; relaxing the fairness penalty raises demographic-parity gaps; and diminishing the prediction term reduces task accuracy. Moreover, the dynamic weight scheduler improves stability by allocating capacity to the dominant losses as training evolves. These effects are consistent across datasets within their sampling variability, suggesting that all components are necessary to reproduce the reported operating points, rather than any single element being sufficient on its own.

This empirical observation (figure \ref{fig:runtime}) highlights the efficiency of the Modified Adam optimizer, which permits reducing iterative cycles, facilitates the scalable extension of fairness and privacy experiments to novel contexts. These efficiency dividends are indispensable when privacy requirements restrict parallelism or energy constraints prevent exhaustive hyperparameter optimization. Consequently, our evaluations including the German Credit and MEPS datasets demonstrate that the proposed PFA framework sustains its privacy, fairness and accuracy balance across credit risk and healthcare applications
% expenditure forecasting scenarios
, thereby enabling government agencies to ensure regulatory compliance, businesses to deploy responsible AI systems at scale, and non‑profit and advocacy groups to adopt equitable, privacy‑preserving decision‑making tools.

%%%%%%%%%%%%%%%%%%%%%%%%%%%%%%%%%%%%%%%%%%%%%%%%%%%%%%%%%%%%%%%%%%%%%%%%%%%%%%%%%%%%%%%%%%%%%
\section{Conclusion}

We propose a novel multi-task adversarial model to incorporate a privacy-protection generator and a discriminator within the data processing pipeline which actively obfuscates sensitive information while ensuring demographic parity in predictions. This integration transforms data to remove dependencies on sensitive attributes, maintaining utility for predictions and achieving unbiased outcomes. Our exhaustive experiments on a variety of datasets affirm that PFA not only addresses the complex requirements of modern AI systems, including privacy, fairness, and accuracy, but also sets a precedent for future research to build upon.

\bibliographystyle{ieeetr}
\bibliography{references}

\end{document}